\documentclass[11pt]{amsart}
\usepackage{preamble}

\begin{document}

\title[Trainable diffusion and density-based losses]{Generative SDEs with Trainable Diffusion and Density-Based Losses}

\begin{abstract}
We develop a divergence-kernel method for jointly optimizing neural SDE drift and diffusion fields and the initial Gaussian distribution, using losses that depend directly on the terminal density.
In two- and eight-dimensional Gaussian-mixture experiments, we learn time-independent fields and obtain far fewer off-mode samples than score-based generators with similar parameter counts.
The drift attracts samples toward the modes, while the diffusion helps disperse excess probability between them.
\end{abstract}

\maketitle

\section{Introduction}

\subsection{Generative models}

Generative models learn a probability distribution from data and provide a way to draw new samples from it.
Diffusion and score-based models form a widely used class of generative models.
They learn time-dependent scores, the spatial gradients of noisy data log-densities, to construct dynamics that reverse a prescribed noising process.
Standard denoising score matching uses explicit conditional scores of Gaussian perturbations as regression targets \cite{Vincent2011,Ho2020,Song2021}.
In these standard constructions, the diffusion coefficient is prescribed rather than jointly optimized with the generative drift.

Flow matching is a closely related approach that learns a time-dependent velocity along a prescribed probability path \cite{Lipman2023}.
It can use the same probability paths as diffusion models but provides conditional velocity targets for regression.
Integrating the resulting ordinary differential equation (ODE) defines a continuous normalizing flow that transports a simple initial distribution toward the data.
The standard formulation has no diffusion field to optimize.
The standard regression objectives in score-based methods and flow matching do not directly minimize losses that depend on the generated terminal density, such as the KL losses used here.

Our aim is to train stochastic differential equation (SDE) generative models with both learnable diffusion and losses that depend directly on the terminal density.
The divergence-kernel (DK) formula provides the density response with respect to both drift and diffusion parameters \cite{DKlinR}.
Learning the diffusion field controls the local strength of stochastic spreading, while the density-based terminal loss determines how distributional errors are penalized.
Both fields can be parameterized to satisfy prescribed bounds or compatible physical structure.

Our experiments use time-independent drift and diffusion, while the same DK formula also supports time-dependent fields.
A fixed dynamical law allows the learned fields to be interpreted and constrained as functions of position.
Time independence may also simplify sampling with fewer time steps, depending on the spatial variation of the learned fields and the numerical solver.

\subsection{Linear-response methods}

Training both fields with density-based losses requires derivatives of the terminal density with respect to the drift and diffusion parameters.
Linear-response methods compute how statistics or probability laws change under parameter perturbations.
Following the review in \cite{DKlinR}, we distinguish path perturbation, divergence formulas, and kernel differentiation methods \cite[Section~5.2]{Ni_kd}.

Path perturbation fixes the Brownian path and differentiates the resulting state path, then averages over realizations.
It includes ensemble sensitivity and adjoint, or backpropagation, methods \cite{eyink2004ruelle}.
This gives derivatives of expectations of smooth observables, but path sensitivities can grow rapidly in unstable systems.
Moreover, these observable-response formulas do not by themselves supply the parameter derivative of the marginal log-density used for KL training.

Divergence formulas differentiate the transfer of probability rather than the state path.
Numerical transfer-operator methods approximate the evolving or invariant density directly \cite{Wormell2019a}, but representing densities becomes costly in high dimension.
For SDEs, differentiating the Fokker--Planck equation gives density-response equations with divergence and second-order terms.
The divergence part of the DK construction represents this density-level response through quantities sampled along trajectories \cite{DKlinR}.

Kernel differentiation keeps the realized state path fixed and differentiates its probability.
This gives likelihood-ratio gradient estimators \cite{Glynn1990}.
Such estimators can have large variance when the noise is small.
In continuous time, the usual Girsanov change of measure handles drift perturbations with fixed diffusion, but does not generally give a finite path-likelihood ratio for a change in the diffusion coefficient \cite{DKlinR}.

Combined response formulas extend the range of dynamics and objectives that can be differentiated.
The fast response formula combines path and divergence contributions for hyperbolic deterministic dynamics \cite{fr}.
The path-kernel method allows responses to perturbations of the initial condition, drift, and diffusion \cite{dud}, and its adjoint form computes parameter gradients of observable objectives \cite{apk}.
The DK method represents the score along sample paths \cite{divKer}; its extension in \cite{DKlinR} gives the marginal-density response for state-dependent and parameter-dependent scalar diffusion.

\subsection{Main results}

Our contribution is a DK training scheme that jointly learns drift and diffusion fields under density-based terminal losses.
We parameterize both fields with neural networks and train the same model using forward KL, backward KL, or their average.
We also optimize the mean and shared standard deviation of the initial Gaussian.
Local parameter derivatives accumulate in one forward pass without storing full trajectories.

The two- and eight-dimensional Gaussian-mixture experiments illustrate both the effect of the loss and the roles of the learned fields.
The three KL objectives lead to different mode-probability and off-mode errors.
The validation-selected models trained with averaged KL produce far fewer off-mode samples than score-based generators of similar parameter count.
The drift points toward the modes, while the diffusion is larger in the gaps, where the initial Gaussian places substantial probability but the target density is low.
Smaller diffusion near the modes limits stochastic spreading there.

This paper is organized as follows.
We present the model and training method in \cref{s:model,s:algorithm}, the numerical comparisons in \cref{s:numerics}, and the discussion and conclusion in \cref{s:discussion}.

\section{Neural SDEs and divergence-kernel gradients}
\label{s:model}

This section specifies the neural SDE model, its terminal losses, and the DK formulas used to compute their parameter gradients.

\subsection{Neural SDE model}

Data preprocessing fixes the spatial scale before training.
For training samples $y_k^{\mathrm{raw}}\in\R^M$, $k=1,\ldots,N_{\mathrm{data}}$, in the original units, set $R=\max_k|y_k^{\mathrm{raw}}|>0$ and $y_k=y_k^{\mathrm{raw}}/R$.
The same fixed $R$ is used for validation and test samples and is independent of $\gamma$.
All subsequent model formulas use normalized coordinates: $x$ denotes the normalized state, and $y_k$ denotes a normalized data point.
Generated samples in the original units are $Rx_T^\gamma$.

The model starts from a Gaussian law and evolves to time $T$.
We use
\begin{equation}
\label{e:sde}
dx_t^\gamma=F^\gamma(x_t^\gamma)\,dt+\sigma^\gamma(x_t^\gamma)\,dB_t,
\qquad x_0\sim N(m_0,s_0^2I_M).
\end{equation}
Here $B_t$ is standard $M$-dimensional Brownian motion, and $\gamma\in\R^{N_\gamma}$ contains all network weights and biases, the initial mean $m_0\in\R^M$, and the shared initial standard deviation $s_0>0$.
The SDE uses the It\^o convention.
Its terminal density is $h_T^\gamma$.
We consider time-independent $F^\gamma$ and $\sigma^\gamma$ throughout this paper, although the theory, including the DK method, also permits time-dependent coefficients.

The initial mean and shared standard deviation can be trained together with the networks.
We optimize $s_0$ directly under the constraint $s_0\ge s_{0,\min}>0$, where $s_{0,\min}$ is a prescribed lower bound in normalized coordinates.
The shared variance is $s_0^2$, and we draw $x_0=m_0+s_0\xi$, where $\xi\sim N(0,I_M)$.
This adds $M+1$ parameters and retains an isotropic initial covariance.
The coordinate standard deviation in the original units is $Rs_0$.

To localize the learned dynamics near the data, we parameterize the drift and diffusion as
\begin{equation}
\label{e:neural_drift}
F^\gamma(x)=\chi(x)\frac{F_{\max}\NN_F(x;\gamma)}{\sqrt{F_{\max}^2+|\NN_F(x;\gamma)|^2}}-(1-\chi(x))F_{\max}\frac{x}{|x|},
\end{equation}
\begin{equation}
\label{e:neural_diffusion}
\sigma^\gamma(x)=\sigma_{\min}+\chi(x)\frac{\sigma_{\max}-\sigma_{\min}}{1+\exp\!\left(-\NN_\sigma(x;\gamma)\right)}.
\end{equation}
The twice continuously differentiable radial cutoff is defined by
\begin{equation}
\label{e:cutoff}
\chi(x)=\begin{cases}
1,&|x|\leq1,\\
1-10s^3+15s^4-6s^5,\quad s=|x|-1,&1<|x|<2,\\
0,&|x|\geq2.
\end{cases}
\end{equation}
Here $\NN$ stands for ``neural network'': $\NN_F$ has $M$ raw outputs, and $\NN_\sigma$ has one raw output before the sigmoid transformation.
The drift cap $F_{\max}>0$ and diffusion bounds $0<\sigma_{\min}<\sigma_{\max}$ are prescribed in normalized units and held fixed during each gradient estimate.
The convex combination in \cref{e:neural_drift} gives $|F^\gamma(x)|\leq F_{\max}$, while \cref{e:neural_diffusion} gives $\sigma_{\min}\leq\sigma^\gamma(x)\leq\sigma_{\max}$.
Inside radius $1$, both fields are determined by bounded transformations of the network outputs.
Between radii $1$ and $2$, the cutoff smoothly blends these outputs with the prescribed tail fields.
At radius $2$ and beyond, the drift points inward with constant magnitude $F_{\max}$, and the diffusion equals its constant floor.
At the origin, the tail term in \cref{e:neural_drift} is defined to be zero by continuous extension.

The networks use smooth activations because the DK updates require second state derivatives.
We use $\operatorname{SiLU}(a)=a/(1+\exp(-a))$ in the hidden layers and take all state derivatives after forming \cref{e:neural_drift,e:neural_diffusion}, including derivatives of $\chi$.
We assume the smoothness, growth control, and integrability required by the DK formulas below.

\subsection{Terminal losses}

We fit the terminal distribution using forward KL, backward KL, or their average.
Let $p$ denote the normalized target density.
Forward KL emphasizes target regions that the model underrepresents ($p\gg h_T^\gamma$), whereas backward KL emphasizes regions where the model assigns excess probability ($p\ll h_T^\gamma$).
We write $\mathcal L$ for the selected terminal KL objective, excluding the weight regularization introduced in \cref{s:parameter_updates}.

Forward KL averages the log density ratio over the target distribution:
\begin{equation}
\label{e:loss}
\mathcal L_{\mathrm{forward}}(\gamma)=D_{\mathrm{KL}}(p\|h_T^\gamma)=\int_{\R^M}p(y)\log\frac{p(y)}{h_T^\gamma(y)}\,dy.
\end{equation}
The training data give the empirical approximation
\[
\mathcal L_{\mathrm{forward}}(\gamma)\approx\frac1{N_{\mathrm{data}}}\sum_{k=1}^{N_{\mathrm{data}}}\log\frac{p(y_k)}{h_T^\gamma(y_k)}.
\]

Backward KL instead averages over the generated distribution:
\[
\mathcal L_{\mathrm{backward}}(\gamma)=D_{\mathrm{KL}}(h_T^\gamma\|p)=\int_{\R^M}h_T^\gamma(x)\log\frac{h_T^\gamma(x)}{p(x)}\,dx.
\]
For $L$ independent generated endpoints $x_{T,l}\sim h_T^\gamma$, $l=1,\ldots,L$, its empirical approximation is
\[
\mathcal L_{\mathrm{backward}}(\gamma)\approx\frac1L\sum_{l=1}^L\log\frac{h_T^\gamma(x_{T,l})}{p(x_{T,l})}.
\]

The averaged KL loss combines these two objectives.
For a fixed target density $p$, with both divergences finite, define
\begin{equation}
\label{e:loss_sym}
\mathcal L_{\mathrm{sym}}(\gamma)=\frac12\mathcal L_{\mathrm{forward}}(\gamma)+\frac12\mathcal L_{\mathrm{backward}}(\gamma).
\end{equation}
This is one half of the Jeffreys divergence, which has also been used to train generative models \cite{Nowozin2016}.
We compare both individual objectives and their average in \cref{s:mixture,s:eightdim}.

Throughout, $\delta$ denotes a parameter derivative at the current $\gamma$, while $\nabla$ denotes a state derivative.
A subscript identifies the parameter: $\delta_{\gamma_i}$ and $\delta_{s_0}$ differentiate with respect to the indicated scalar parameter, and $\delta_{m_0}$ collects the mean-component derivatives into a column vector.
In particular, $\delta F^\gamma$ and $\delta\sigma^\gamma$ hold the state fixed.
The log-density parameter derivatives below are also evaluated at fixed spatial arguments.

\subsection{Score computed by the forward process}

The DK formulas express density derivatives as conditional means of pathwise terms.
We recall the formulas from \cite{divKer,DKlinR} for scalar diffusion.
Numerical checks of the response formula for a one-dimensional multiplicative-noise SDE and a $40$-dimensional Lorenz~96 observable are given in \cite[Section~4]{DKlinR}.
The exact discrete score and response identities are given in \cite[Section~3.2]{DKlinR}.
The continuous formulas below are the formal limits used to build the time-step scheme.

The covector $\nu_t\in\R^M$ carries the score information along each path.
We use the general DK score formula of \cite{DKlinR} with a scalar backward-adapted transfer rate $\alpha_t$, which may depend on the path from $t$ to $T$.
The stochastic expressions below are interpreted as limits of \cref{e:nu_step,e:response_step}, with right-endpoint evaluation of $\alpha_{n+1}$ and start-of-step evaluation of the other coefficients.
The drift, diffusion, and their state derivatives without the superscript $\gamma$ are evaluated at the current parameter and at $x_t$.
The covector starts at
\[
\nu_0=\nabla\log h_0^\gamma(x_0)=-\frac{x_0-m_0}{s_0^2}.
\]
It evolves by
\begin{equation}
\label{e:nu_sde}
\begin{aligned}
d\nu_t={}&\left[
(\nabla\sigma\nabla\sigma^T-\nabla F^T-\alpha_t I_M)\nu_t
-\nabla\div F+\nabla^2\sigma\nabla\sigma+\nabla\sigma\,\Delta\sigma
\right]dt\\
&-\left[\nabla\sigma\,\nu_t^T+\nabla^2\sigma+\frac{\alpha_t}{\sigma}I_M\right]dB_t.
\end{aligned}
\end{equation}
Here $\nabla F$ is the state Jacobian, $\nabla^2\sigma$ is the Hessian, and $\Delta\sigma=\sum_{i=1}^M\partial_i^2\sigma$ is the Laplacian.
Throughout, $\E$ denotes expectation over paths of the SDE under consideration or its time-discretized simulation.
Target-data averages are written explicitly as integrals or empirical sums.
The score identity is
\[
\nabla\log h_t(x_t)=\E[\nu_t\mid x_t].
\]
Thus one covector, with $M$ entries, serves all parameter derivatives.

\subsection{Response of the terminal density and loss}

We first express the terminal-density response as a conditional expectation of a pathwise integral, then use this identity to derive the responses of the forward, backward, and averaged KL losses.
For a parameter derivative, let $\mathcal T_T$ be the following pathwise response:
\begin{equation}
\label{e:response_sde}
\begin{aligned}
\mathcal T_T={}&\delta\log h_0^\gamma(x_0)
+\int_0^T\left[
\delta\sigma^\gamma\,\nabla\sigma^T\nu_t
-\delta F^{\gamma T}\nu_t-\div\delta F^\gamma
+\delta\sigma^\gamma\,\Delta\sigma
+\nabla\sigma^T\nabla\delta\sigma^\gamma
\right]dt\\
&-\int_0^T\left[\delta\sigma^\gamma\,\nu_t^T+
(\nabla\delta\sigma^\gamma)^T\right]dB_t.
\end{aligned}
\end{equation}
Here, the initial-density term accounts for changes in the Gaussian mean and variance.
Differentiating its log density at fixed $x_0$ gives
\begin{equation}
\label{e:initial_response}
\mathcal T_0=\delta\log h_0^\gamma(x_0)=\frac{(x_0-m_0)^T\delta m_0}{s_0^2}+\left(\frac{|x_0-m_0|^2}{s_0^2}-M\right)\frac{\delta s_0}{s_0}.
\end{equation}
This term vanishes for derivatives with respect to the network parameters.

The DK identity recovers the terminal log-density derivative by conditioning the pathwise response on the endpoint.
\begin{equation}
\label{e:conditional}
\delta\log h_T^\gamma(x)=\E[\mathcal T_T\mid x_T=x].
\end{equation}
Since the terminal density integrates to one, taking expectations in \cref{e:conditional} gives
\[
\E[\mathcal T_T]=\E\!\left[\E[\mathcal T_T\mid x_T]\right]=\int_{\R^M}h_T^\gamma(x)\,\delta\log h_T^\gamma(x)\,dx=\delta\int_{\R^M}h_T^\gamma(x)\,dx=\delta 1=0.
\]

For forward KL in \cref{e:loss}, the target density is fixed during differentiation.
Using \cref{e:conditional}, we obtain
\[
\begin{aligned}
\delta\mathcal L_{\mathrm{forward}}
&=-\int_{\R^M}p(y)\,\delta\log h_T^\gamma(y)\,dy=-\int_{\R^M}p(y)\,\E[\mathcal T_T\mid x_T=y]\,dy\\
&=-\int_{\R^M}h_T^\gamma(x)\frac{p(x)}{h_T^\gamma(x)}\,\E[\mathcal T_T\mid x_T=x]\,dx=-\E\!\left[\frac{p(x_T)}{h_T^\gamma(x_T)}\mathcal T_T\right].
\end{aligned}
\]
Here, multiplying and dividing by $h_T^\gamma(x)$ rewrites the target average in terms of generated samples.
The final equality uses the law of total expectation.
Using $L$ generated SDE paths, we approximate the forward-KL response by
\[
\delta\mathcal L_{\mathrm{forward}}\approx-\frac1L\sum_{l=1}^L\frac{p(x_{T,l})}{h_T^\gamma(x_{T,l})}\,\mathcal T_{T,l},
\]
where $\mathcal T_{T,l}$ is the response accumulated along path $l$.
Here, the factor $p(x_{T,l})/h_T^\gamma(x_{T,l})$ is the terminal coefficient for forward KL.
At optimization update $m$, we denote the estimated terminal coefficient for generated path $l$ by $a_l^{(m)}$.
\Cref{e:terminal_coefficients} specifies it for each KL objective.

Backward KL also differentiates the generated density that weights the integral:
\[
\delta\mathcal L_{\mathrm{backward}}=\int_{\R^M}h_T^\gamma(x)\left[1+\log\frac{h_T^\gamma(x)}{p(x)}\right]\delta\log h_T^\gamma(x)\,dx.
\]
Using \cref{e:conditional} and $\E[\mathcal T_T]=0$, the constant term vanishes, giving
\[
\begin{aligned}
\delta\mathcal L_{\mathrm{backward}}
&=\int_{\R^M}h_T^\gamma(x)\log\frac{h_T^\gamma(x)}{p(x)}\,\delta\log h_T^\gamma(x)\,dx\\
&=\int_{\R^M}h_T^\gamma(x)\log\frac{h_T^\gamma(x)}{p(x)}\,\E[\mathcal T_T\mid x_T=x]\,dx\\
&=\E\!\left[\log\frac{h_T^\gamma(x_T)}{p(x_T)}\mathcal T_T\right]=-\E\!\left[\log\frac{p(x_T)}{h_T^\gamma(x_T)}\mathcal T_T\right].
\end{aligned}
\]
The third line again uses the law of total expectation.
For the generated endpoints, the corresponding empirical approximation is
\[
\delta\mathcal L_{\mathrm{backward}}\approx\frac1L\sum_{l=1}^L\log\frac{h_T^\gamma(x_{T,l})}{p(x_{T,l})}\,\mathcal T_{T,l}.
\]

The averaged KL derivative is the average of the forward and backward derivatives, so
\[
\delta\mathcal L_{\mathrm{sym}}=\frac12\delta\mathcal L_{\mathrm{forward}}+\frac12\delta\mathcal L_{\mathrm{backward}}=-\frac12\E\!\left[\left(\frac{p(x_T)}{h_T^\gamma(x_T)}+\log\frac{p(x_T)}{h_T^\gamma(x_T)}\right)\mathcal T_T\right].
\]
The coefficients in these response formulas are evaluated at the current parameters and held fixed when computing the local response.
The discrete counterparts of \cref{e:nu_sde,e:response_sde,e:conditional} are given in \cref{s:algorithm} and provide the gradient used for training.

Subtracting a fixed baseline from the coefficient leaves the exact gradient unchanged because $\E[\mathcal T_T]=0$.
At optimization update $m$, let $b_m$ be a scalar determined before drawing the current paths. Then
\[
\delta\mathcal L_{\mathrm{sym}}=-\E\!\left[\left\{\frac12\left(\frac{p(x_T)}{h_T^\gamma(x_T)}+\log\frac{p(x_T)}{h_T^\gamma(x_T)}\right)-b_m\right\}\mathcal T_T\right].
\]
We can take $b_m$ to be the mean of the coefficients from a previous update.
This is a simple choice for reducing the variance of the estimator, although it does not generally minimize the variance.
The detailed formula for $b_m$ is given in \cref{s:terminal_weights}.

\section{Training algorithm}
\label{s:algorithm}

Training combines forward response accumulation with terminal weights estimated from samples.
We keep the model parameters fixed throughout each gradient estimate.
The state, covector, and response use the same Brownian increments.

\subsection{State, covector, and response steps}

The state follows the Euler scheme for \cref{e:sde}.
Set $t_n=n\Dt$, $N\Dt=T$, and $\DB_n=B_{t_{n+1}}-B_{t_n}$.
Then
\[
x_{n+1}=x_n+F_n\Dt+\sigma_n\DB_n,
\qquad \DB_n\sim N(0,\Dt I_M).
\]
The increments are independent over steps and paths.
Write $F_n=F^\gamma(x_n)$ and $\sigma_n=\sigma^\gamma(x_n)$; their state derivatives are also evaluated at $x_n$.

The covector step evaluates the transfer rate $\alpha$ at the new state and all other coefficients at the start of the step.
With discrete transfer weight $\alpha_{n+1}\Dt$, the small-step expansion of the score recursion in \cite[Sections~3.2--3.3]{DKlinR} is
\begin{equation}
\label{e:nu_step}
\begin{aligned}
\nu_{n+1}={}&\nu_n+\left[
(\nabla\sigma_n\nabla\sigma_n^T-\nabla F_n^T-\alpha_{n+1}I_M)\nu_n
-\nabla\div F_n+\nabla^2\sigma_n\nabla\sigma_n
+\nabla\sigma_n\,\Delta\sigma_n
\right]\Dt\\
&-\left[\nabla\sigma_n\nu_n^T+\nabla^2\sigma_n+\frac{\alpha_{n+1}}{\sigma_n}I_M\right]\DB_n.
\end{aligned}
\end{equation}
The results of \cite{DKlinR} allow $\alpha_{n+1}$ to depend on $x_{n+1}$.
We compute $x_{n+1}$ before evaluating $\alpha_{n+1}$, so this choice still permits forward simulation.

We choose $\alpha$ proportional to $\sigma$ to balance covector damping against stochastic forcing in \cref{e:nu_step}.
If $\alpha$ is too small, damping may be insufficient.
If it is too large, the coefficient $\alpha_{n+1}/\sigma_n$ can amplify the noise and increase fluctuations in $\nu$.
Scaling $\alpha$ with the local diffusion helps control this coefficient, while prescribing the mean of $\alpha$ sets the overall damping strength.
The ratio $\alpha_{n+1}/\sigma_n$ in the kernel-differentiation term is generally not constant, because $\alpha_{n+1}$ is evaluated at $x_{n+1}$ while $\sigma_n$ is evaluated at $x_n$.
For a prescribed mean-rate target $\alpha_{\mathrm{target}}>0$, set
\begin{equation}
\label{e:alpha_mean}
\alpha_{n+1}=\frac{\alpha_{\mathrm{target}}}{\overline\sigma_{\mathrm{prev}}}\sigma^\gamma(x_{n+1}),
\end{equation}
where $\overline\sigma_{\mathrm{prev}}$ is the mean endpoint diffusion over all paths and time steps in the previous update.

The target $\alpha_{\mathrm{target}}$ specifies the desired average of $\alpha$ over all paths and time steps.
This average remains close to the target when the current mean diffusion is close to $\overline\sigma_{\mathrm{prev}}$.
Individual values of $\alpha$ still vary with the local diffusion and need not be close to $\alpha_{\mathrm{target}}$.
For the first update, $\overline\sigma_{\mathrm{prev}}$ is evaluated on an independent batch from the initial Gaussian.
With $\overline\sigma_{\mathrm{prev}}$ fixed, $\alpha_{n+1}$ depends only on the endpoint state and satisfies the backward-adapted requirement.

The response is accumulated at each time step using the covector $\nu_n$ at the start of the step.
Starting from $\mathcal T_0$ in \cref{e:initial_response}, the time-step form of \cref{e:response_sde} is
\begin{equation}
\label{e:response_step}
\begin{aligned}
\mathcal T_{n+1}-\mathcal T_n={}&
\nu_n^T\left(-\delta F_n^\gamma\Dt-\delta\sigma_n^\gamma\DB_n
+\delta\sigma_n^\gamma\nabla\sigma_n\Dt\right)\\
&+\delta\sigma_n^\gamma\Delta\sigma_n\Dt
-\div\delta F_n^\gamma\Dt
-(\nabla\delta\sigma_n^\gamma)^T\DB_n
+\nabla\sigma_n^T\nabla\delta\sigma_n^\gamma\Dt.
\end{aligned}
\end{equation}
All fields in this expression use $(t_n,x_n)$.
The response increment is accumulated before replacing $\nu_n$ by $\nu_{n+1}$.
\Cref{e:nu_step,e:response_step} are time-step approximations of the exact discrete DK identities in \cite[Sections~3.2--3.4]{DKlinR}, obtained using the It\^o quadratic-variation rule in the small-step limit.
If $h_N^\gamma$ is the density of the Euler terminal state, they give
\[
\delta\log h_N^\gamma(x)\approx\E[\mathcal T_N\mid x_N=x].
\]

Local automatic differentiation computes the response increments from a scalar function.
For this calculation, $x_n$, $\nu_n$, $\DB_n$, and all coefficients without the superscript $\gamma$ are held fixed at their current values.
Define
\begin{equation}
\label{e:local}
\begin{aligned}
\ell_n^\gamma={}&-\Dt\left[\nu_n^TF_n^{\gamma}+\div F_n^{\gamma}\right]\\
&+\sigma_n^{\gamma}\left[\Dt\left(\nabla\sigma_n^T\nu_n+\Delta\sigma_n\right)-\nu_n^T\DB_n\right]
+(\nabla\sigma_n^{\gamma})^T\left(\Dt\nabla\sigma_n-\DB_n\right).
\end{aligned}
\end{equation}
In the code, we first form $\ell_n^\gamma$ in \cref{e:local}, then obtain the response increment $\delta\ell_n^\gamma=\mathcal T_{n+1}-\mathcal T_n$ by automatic differentiation.

\subsection{Terminal weights and density estimates}
\label{s:terminal_weights}

Gaussian kernel density estimates (KDEs) supply the terminal KL coefficients from $L$ generated endpoints and the training data.
For bandwidth $\tau>0$, let $K_\tau(z)=(2\pi\tau^2)^{-M/2}\exp(-|z|^2/(2\tau^2))$.
At a generated endpoint $x_{N,l}$, the estimates are
\[
\widehat p(x_{N,l})=\frac1{N_{\mathrm{data}}}\sum_{k=1}^{N_{\mathrm{data}}}K_\tau(x_{N,l}-y_k),
\qquad
\widehat h_N^\gamma(x_{N,l})=\frac1L\left[\frac12K_\tau(0)+\sum_{j\ne l}K_\tau(x_{N,l}-x_{N,j})\right].
\]
Every other sample contributes its full kernel value, the query's own contribution has weight $1/2$, and the denominator remains the full population size.

The same response accumulation supports all three KL objectives.
At optimization update $m=0,1,\ldots$, the coefficient for path $l$ is
\begin{equation}
\label{e:terminal_coefficients}
a_l^{(m)}=\begin{cases}
\widehat p(x_{N,l})/\widehat h_N^\gamma(x_{N,l}),&\text{forward KL},\\
\log\!\left(\widehat p(x_{N,l})/\widehat h_N^\gamma(x_{N,l})\right),&\text{backward KL},\\
\tfrac12\left[\widehat p(x_{N,l})/\widehat h_N^\gamma(x_{N,l})+\log\!\left(\widehat p(x_{N,l})/\widehat h_N^\gamma(x_{N,l})\right)\right],&\text{averaged KL}.
\end{cases}
\end{equation}
The endpoints, parameters, and density estimates are those of update $m$.

The previous update's mean coefficient provides a baseline for the current gradient estimate.
We subtract $b_m$ from $a_l^{(m)}$ in \cref{e:gradient} to reduce the variance of the gradient estimator.
We use
\[
b_0=0,\qquad b_{m+1}=\frac1L\sum_{l=1}^L a_l^{(m)}.
\]
Fresh paths are drawn at each update, with $b_m$ frozen before drawing them.
The same baseline is used for the drift, diffusion, and initial-law parameters; its new value is retained only after an accepted parameter update.

The DK gradient formula does not require the scalar KL loss value, but we compute its KDE estimate, denoted by $\widehat{\mathcal L}$, to monitor optimization and set the clipping budget in \cref{s:parameter_updates}.
We estimate forward KL by averaging over data points and backward KL by averaging over generated endpoints; averaged KL is the mean of these two estimates.
For forward and averaged KL, we also evaluate the densities at each data point $y_k$, where the same half-weight rule applies to its own contribution to $\widehat p(y_k)$:
\[
\widehat p(y_k)=\frac1{N_{\mathrm{data}}}\left[\frac12K_\tau(0)+\sum_{i\ne k}K_\tau(y_k-y_i)\right],
\qquad
\widehat h_N^\gamma(y_k)=\frac1L\sum_{l=1}^L K_\tau(y_k-x_{N,l}).
\]

\subsection{Gradient estimation in one forward pass}

We estimate loss derivatives for both the neural networks and the initial Gaussian using the same terminal weights.
For a parameter derivative $\delta$ of a network weight or bias, let $\mathcal T_{N,l}$ be the accumulated response for path $l$.
At optimization update $m$, using the coefficients $a_l^{(m)}$ and baseline $b_m$ from \cref{s:terminal_weights}, we get
\begin{equation}
\label{e:gradient}
\widehat{\delta\mathcal L}=-\frac1L\sum_{l=1}^L(a_l^{(m)}-b_m)\mathcal T_{N,l}.
\end{equation}
This estimates the KL parameter derivative using the DK response and estimated density ratios; it need not equal the derivative of the scalar KDE estimate $\widehat{\mathcal L}$.

The initial Gaussian gives explicit gradients for its mean and standard deviation.
The initial-law parameters are separate from the network weights, so their response is therefore given entirely by \cref{e:initial_response}, with no further time accumulation.
Using the coefficient for the selected KL objective from \cref{s:terminal_weights}, the initial-law gradients are
\begin{equation}
\label{e:initial_gradients}
\begin{gathered}
\widehat{\delta_{m_0}\mathcal L}=-\frac1L\sum_{l=1}^L(a_l^{(m)}-b_m)\frac{x_{0,l}-m_0}{s_0^2},
\\
\widehat{\delta_{s_0}\mathcal L}=-\frac1{Ls_0}\sum_{l=1}^L(a_l^{(m)}-b_m)\left(\frac{|x_{0,l}-m_0|^2}{s_0^2}-M\right).
\end{gathered}
\end{equation}
The parameter derivatives $\delta_{m_0}\log h_0^\gamma(x_0)$ and $\delta_{s_0}\log h_0^\gamma(x_0)$ are read from the $\delta m_0$ and $\delta s_0$ terms in \cref{e:initial_response}, respectively.
They have zero mean under the initial Gaussian, so the same baseline subtraction applies.
In these derivatives, $x_{0,l}$ is held fixed once it was sampled as $m_0+s_0\xi_l$.

All parameter gradients are computed in one forward pass.
At the initial sample, we store the derivatives of $\log h_0^\gamma(x_{0,l})$ with respect to $m_0$ and $s_0$.
During simulation, we accumulate the network-parameter responses using \cref{e:response_step,e:local}, retaining only the current state, covector, and accumulated responses for each path.
After all paths finish, we compute the terminal coefficients and obtain the network and initial-law gradients using \cref{e:gradient,e:initial_gradients}, respectively.

The memory cost is independent of the number of time steps.
The accumulated responses require $O(LN_\gamma)$ entries, and full trajectories are not stored.

\subsection{Parameter updates and constraints}
\label{s:parameter_updates}

The regularized training objective adds a weight penalty to the selected terminal KL.
Let $W$ range over all weight matrices in the drift and diffusion networks, and let $\|\cdot\|_F$ denote the Frobenius norm.
The objective is
\[
\mathcal L_{\mathrm{reg}}(\gamma)=\mathcal L(\gamma)+\frac{10^{-5}}2\sum_W\|W\|_F^2.
\]
Biases and initial-law parameters are excluded from this penalty.
We add its exact derivative to the KL derivative estimate in \cref{e:gradient,e:initial_gradients}:
\[
\widehat{\delta\mathcal L_{\mathrm{reg}}}=\widehat{\delta\mathcal L}+\delta\left(\frac{10^{-5}}2\sum_W\|W\|_F^2\right).
\]

The optimizer forms a proposed increment from these regularized gradient estimates.
At update $m$, all derivatives are evaluated at $\gamma^{(m)}$, and $\Delta\gamma\in\R^{N_\gamma}$ denotes an increment of the full parameter vector, including $m_0$ and $s_0$.
For stochastic gradient descent with step size $\eta_m>0$, its components are
\[
\Delta\gamma_i=-\eta_m\widehat{\delta_{\gamma_i}\mathcal L_{\mathrm{reg}}}.
\]
The reported experiments use Adam to form this increment from the same regularized gradient estimates.

We enforce the lower bound on the proposed initial standard deviation before clipping.
Both numerical examples use $s_{0,\min}=0.3$.
Writing $\Delta s_0$ for the $s_0$ component of $\Delta\gamma$, we replace it with
\[
\Delta s_0\leftarrow\max\{\Delta s_0,\;s_{0,\min}-s_0\}.
\]
Also note that the parameterizations in \cref{e:neural_drift,e:neural_diffusion} enforce $|F^\gamma(x)|\leq F_{\max}$ and $\sigma_{\min}\leq\sigma^\gamma(x)\leq\sigma_{\max}$.
These field bounds require no additional projection of the network parameters.

We clip the proposed increments separately for the drift network, the diffusion network, and the initial Gaussian.
Each network group contains its weights and biases, while the initial-Gaussian group contains the $M$ components of $m_0$ and the shared standard deviation $s_0$.
For a group with parameter indices $G$, we limit the magnitude of its estimated first-order change in the unregularized KL loss to $|\widehat{\mathcal L}|/30$:
\[
\left|\sum_{i\in G}\widehat{\delta_{\gamma_i}\mathcal L}\,\Delta\gamma_i\right|\leq\frac1{30}|\widehat{\mathcal L}|.
\]
If this bound is exceeded, we multiply all increments in the group by the ratio of the allowed magnitude to the predicted magnitude.
This common scaling factor lies in $[0,1]$, so the previously enforced lower bound on $s_0$ is preserved.

After projection and clipping, we apply the resulting increment:
\[
\gamma^{(m+1)}=\gamma^{(m)}+\Delta\gamma.
\]

\section{Numerical study}
\label{s:numerics}

We study two- and eight-dimensional four-component Gaussian mixtures.
In each dimension, we compare DK models trained with forward, backward, and averaged KL against an EDM baseline, using identical data sets.
EDM is a score-based diffusion framework that achieved state-of-the-art image-generation results \cite{Karras2022}.
Both methods use plain multilayer perceptrons (MLPs), with a slightly larger parameter count for EDM.
The DK models produce fewer samples outside the data neighborhoods, and these outside samples have smaller mean distances to the nearest reference points.

\subsection{Evaluation protocol}
\label{s:density_evaluation}

In each experiment, all methods use identical training, validation, and test sets and the same evaluation protocol.
We draw independent training, validation, and test sets containing $1{,}024$, $1{,}024$, and $10{,}000$ points, respectively.
For each experiment, we compute $R$ from the training set and use it to normalize the training, validation, and test sets, as in \cref{s:model}.

KL evaluation uses Gaussian kernel density estimates computed from all reference data points and generated samples, as described in \cref{s:terminal_weights}.
These estimates use half-weight self contributions and division by the total number of points in the corresponding set.
Forward KL averages $\log(\widehat p/\widehat h_N^\gamma)$ over reference points, and backward KL averages $\log(\widehat h_N^\gamma/\widehat p)$ over generated points.
We report both KLs and their average, called averaged KL.
For validation and testing, we fix the Gaussian kernel bandwidth defined in \cref{s:terminal_weights} at $\tau=0.03$, independently of the training bandwidth.

Checkpoint selection uses validation averaged KL for every training objective.
Every $50$ updates, each method generates $1{,}024$ samples and compares them with the shared validation set.
We select the checkpoint with lowest validation averaged KL and evaluate it using $10{,}000$ generated samples and the shared test set.

Sample neighborhoods measure how often generated points lie away from the observed distribution.
Following the precision construction of \cite{Kynkaanniemi2019}, we center a closed ball at each point in the target test set, with radius equal to its distance to the third nearest other reference point.
Precision is the fraction of generated samples in the union of these balls; the outside fraction is its complement.
For the outside samples, we also report the mean and $95$th percentile of the distance to the nearest reference point, measured in normalized coordinates.
These distances are to the reference points, rather than to the boundaries of the balls.
Neither the neighborhoods nor these distances require the analytic target density.

We also compare the probability in the four regions associated with the Gaussian components.
We assign each generated sample and target test point to the nearest component mean, then compute the fraction of points in each region.
We report the $L^1$ difference between the generated and target region probabilities.
This diagnostic uses the known component means, whereas the neighborhood metrics use only samples.

We use two independent sample sets drawn directly from the target distribution as a common reference for DK and EDM.
Each set contains $10{,}000$ points, and we apply the same evaluation protocol.
Since both sets follow the same target distribution, the population KL is zero; any nonzero estimated KL reflects finite-sample density-estimation error.

We measure training-run wall time and network-only forward time on NVIDIA L40S GPUs.
Wall times include compilation, validation, testing, plotting, and checkpoint writing.
Forward times are median PyTorch CUDA-graph measurements for batches of $1{,}024$ float32 inputs, with the two DK MLPs evaluated on concurrent CUDA streams; compilation, host transfers, and operations outside the MLPs are excluded.

\subsection{Two-dimensional experiments}
\label{s:mixture}

The target has four components with different probabilities, widths, and orientations.
In the original data units, its density is
\[
p_{\mathrm{raw}}(y^{\mathrm{raw}})=\sum_{j=0}^{3}\pi_j\varphi_2(y^{\mathrm{raw}};\mu_j,\Sigma_j),
\]
where $\varphi_2(\cdot;\mu,\Sigma)$ is the density of $N(\mu,\Sigma)$.
The component weights and means are
\[
\pi_j=\frac{j+1}{10},\qquad
\theta_j=\frac{2\pi j}{4},\qquad
\mu_j=3(\cos\theta_j,\sin\theta_j)^T.
\]
The covariance matrices are
\[
\Sigma_j=Q(\theta_j+\pi/6)\diag(a_j^2,b_j^2)Q(\theta_j+\pi/6)^T,
\]
where $a_j=0.35+0.05j$ and $b_j=0.15+0.02j$ are the standard deviations along the principal axes, and
\[
Q(\theta)=\begin{pmatrix}\cos\theta&-\sin\theta\\\sin\theta&\cos\theta\end{pmatrix}
\]
is the planar rotation matrix.

We rescale the data by the training-derived radius $R$ specified in \cref{s:density_evaluation}.
Conditional on this fixed $R$, the normalized target density is $p(y)=R^2p_{\mathrm{raw}}(Ry)$, with component means $\mu_j/R$ and covariances $\Sigma_j/R^2$.
Training and the density-based evaluation metrics use only samples, rather than the analytic mixture density.

\begingroup
\renewcommand{\floatpagefraction}{0.7}
\begin{table}[p]
\centering
\caption{Two-dimensional validation and test results for checkpoints selected by minimum validation averaged KL during $5{,}000$ training updates.
Validation uses $1{,}024$ generated and $1{,}024$ reference samples; test metrics use $10{,}000$ of each.
KL uses all-point Gaussian estimates with bandwidth $0.03$ and half-weight self contributions.
Outside-sample distances are in normalized coordinates.}
\label{t:twodim}
\small
\setlength{\tabcolsep}{4pt}
\begin{tabular}{lrrrr}
\toprule
 & \multicolumn{3}{c}{DK training objective} & \\
\cmidrule(lr){2-4}
Metric & Averaged KL & Forward KL & Backward KL & EDM \\
\midrule
Selected update & $4700$ & $4900$ & $4500$ & $4950$ \\
Parameter count & $1740$ & $1740$ & $1740$ & $1827$ \\
Validation averaged KL & $0.103$ & $0.0892$ & $0.0948$ & $0.180$ \\
Test averaged KL & $0.0690$ & $0.0882$ & $0.0804$ & $0.118$ \\
Test forward KL & $0.0360$ & $0.0358$ & $0.0506$ & $0.0475$ \\
Test backward KL & $0.102$ & $0.141$ & $0.110$ & $0.189$ \\
Precision & $98.2\%$ & $96.9\%$ & $98.1\%$ & $96.3\%$ \\
Outside fraction & $1.80\%$ & $3.06\%$ & $1.92\%$ & $3.69\%$ \\
Mean outside distance & $0.0209$ & $0.0459$ & $0.0124$ & $0.0545$ \\
Outside distance: $95$th percentile & $0.0802$ & $0.198$ & $0.0490$ & $0.197$ \\
Mode-probability $L^1$ error & $0.0552$ & $0.0620$ & $0.0742$ & $0.0742$ \\
\bottomrule
\end{tabular}
\end{table}

\begin{figure}[p]
\centering
\includegraphics[width=\textwidth]{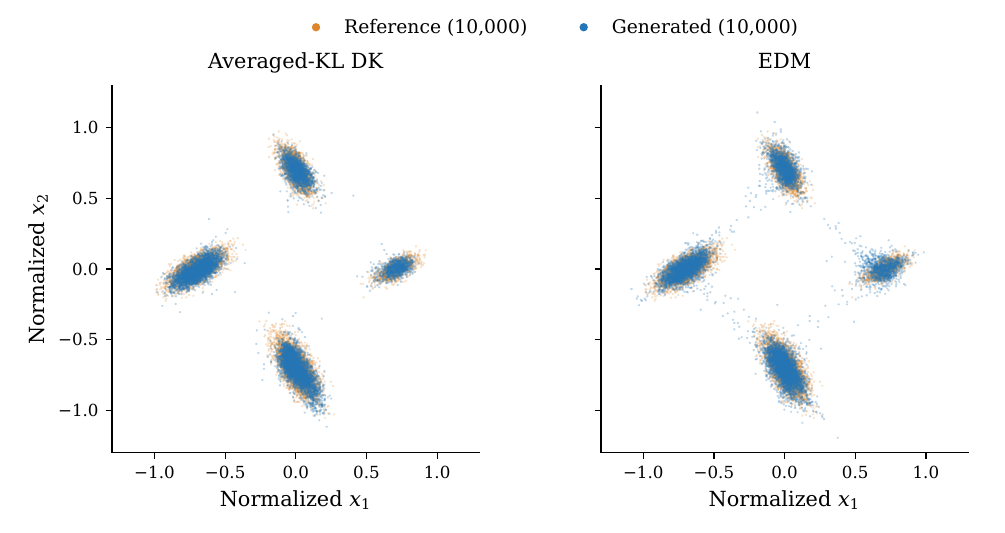}
\caption{Two-dimensional test samples from the minimum-validation-KL checkpoints: averaged-KL DK at update $4{,}700$ (left) and plain-MLP EDM at update $4{,}950$ (right).
Each panel shows all $10{,}000$ generated samples in blue and the same $10{,}000$ independent reference samples in orange, using normalized coordinates, identical axis limits, and equal aspect ratios.}
\label{f:twodim_samples}
\end{figure}

DK jointly trains two networks for the time-independent drift and diffusion, together with the initial Gaussian law, for a total of $1{,}740$ parameters.
The drift network has three hidden layers of width $24$ and two outputs, and the diffusion network has two hidden layers of width $18$ and one output.
Both use SiLU activations and take only the normalized location as input.
We initialize the hidden layers sequentially using the normalized training data.
For each layer, we rescale its random weight matrix so that the preactivation has root-mean-square value one across the training points and hidden units, then apply SiLU to produce the next layer's inputs.
Hidden biases are zero, and the final layers have small random weights and zero biases.
The initial mean and standard deviation are initialized at $m_0=0$ and $s_0=0.9/\sqrt2$.
Each update simulates $L=1{,}024$ paths to $T=1$ with $\Dt=0.002$, or $500$ time steps.
The fields use \cref{e:neural_drift,e:neural_diffusion} with drift cap $4$ and diffusion bounds $0.1$ and $2$.

The training density bandwidth decreases linearly from $0.3$ at update $1$ to $0.03$ at update $4{,}000$, then remains fixed through update $5{,}000$.
The drift and diffusion bounds remain fixed throughout training.
The transfer rate follows \cref{e:alpha_mean} with $\alpha_{\mathrm{target}}=42$.

The three DK runs differ in their terminal objective: forward, backward, or averaged KL.
They use the coefficients in \cref{e:terminal_coefficients} with the previous-update mean subtracted and the update rules in \cref{s:parameter_updates}.
All four runs, including EDM, use Adam with learning rate $10^{-3}$, batch size $1{,}024$, and $5{,}000$ updates, with training seed $0$.

The EDM baseline uses a noise-dependent plain MLP with $1{,}827$ trainable parameters, slightly more than the two DK networks combined.
It concatenates the preconditioned location and $16$ fixed sine and cosine noise features, followed by three SiLU hidden layers of width $25$ and a linear two-component output.
The MLP has no internal residual connections or layer normalization; the outer EDM preconditioning is retained.
Hidden and final-layer weights use Glorot initialization, with final-layer weights additionally scaled by $10^{-5}$; all biases are zero.

EDM training follows the preconditioned denoising objective of \cite{Karras2022}.
We apply a fixed rescaling fitted on the training data so that the root-mean-square coordinate standard deviation is $0.5$, and undo this rescaling before evaluation and plotting.
Training noise levels are lognormally distributed with log-mean $-1.2$ and log-standard deviation $1.2$ in these internal coordinates.
Generation uses the exponential moving average of the weights with decay $0.99$ and $128$ deterministic Heun steps, with noise levels decreasing from $80$ to $0.002$ using the EDM schedule exponent $7$.
The final step is Euler to zero noise, giving $255$ network evaluations per generated sample.
The EDM gradient norm is not clipped.

The EDM baseline retains its preconditioning, exponential moving average of the weights, and second-order Heun sampler, and uses slightly more trainable parameters.
DK uses two smaller networks for the drift and diffusion, whose evaluations could be performed in parallel.

In the two-dimensional experiment, DK produces fewer off-mode samples than EDM, and the remaining off-mode samples lie closer to the reference data on average (\cref{t:twodim}).
\Cref{f:twodim_samples} makes the stronger concentration of DK samples near the modes visible.
Among the DK objectives, backward-KL training gives the smallest mean outside distance, while averaged-KL training gives the lowest mode-probability error.
All three DK models have lower validation and test averaged KL than EDM, despite DK's less extensively engineered and tuned implementation and smaller parameter count.
Averaged-KL training gives the lowest test averaged KL.

As a common reference for DK and EDM, we compare two independent sample sets drawn directly from the target distribution.
This gives estimated averaged KL $0.00415$ and precision $98.4\%$.
The averaged-KL DK model nearly matches this reference in precision.

The learned drift and diffusion play complementary roles.
\Cref{f:learned_fields} shows the drift pointing toward the modes, qualitatively resembling a score field.
The diffusion is largest in the central region, where the initial Gaussian places substantial probability but the target density is low, and is close to its floor elsewhere.
This pattern can help disperse probability from the central gap while limiting stochastic spreading.
This suggests combining score-based generation with DK refinement.

\begin{figure}[tbp]
\centering
\includegraphics[width=\textwidth]{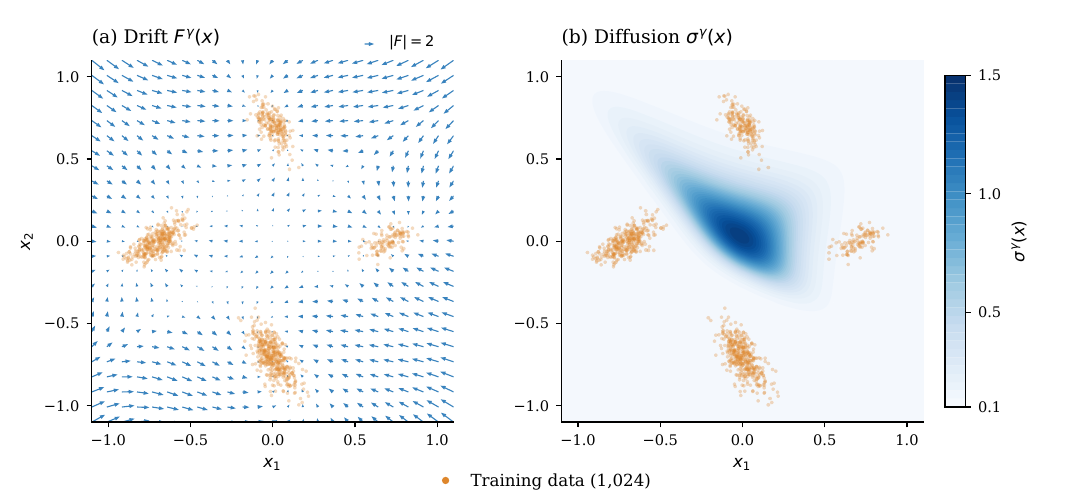}
\caption{Learned two-dimensional fields from the averaged-KL DK checkpoint selected at update $4{,}700$.
Left: the drift $F^\gamma$, with arrows anchored on a uniform grid.
Right: filled contours of the scalar diffusion $\sigma^\gamma$.
Both panels use the same normalized coordinates and show the $1{,}024$ training points in orange.}
\label{f:learned_fields}
\end{figure}

The $5{,}000$-update averaged-KL DK and EDM runs take $924$ and $159$ seconds, respectively.
Their median network-only forward times are both $27.9$ microseconds per batch.
\endgroup

\subsection{Eight-dimensional experiments}
\label{s:eightdim}

We first generate an eight-dimensional Gaussian-mixture vector $U=(U_1,\ldots,U_8)^\top$ in the original data units, using the same component weights as the two-dimensional target.
Within component $j$, the first two coordinates $U_1,U_2$ have mean $\mu_j$ and covariance $\Sigma_j$.
We add six jointly Gaussian coordinates $U_3,\ldots,U_8$, with zero mean and covariance
\[
\operatorname{Cov}(U_a,U_b)=\frac{\tr(\Sigma_j)}{2}(0.2+0.1j)^{|a-b|},\qquad a,b=3,\ldots,8.
\]
Within each component, these six coordinates are independent of the first two, so
\[
\operatorname{Cov}(U_a,U_b)=0,\qquad a=1,2,\quad b=3,\ldots,8.
\]
That is, each added coordinate has variance $\tr(\Sigma_j)/2$, while correlations among the added coordinates decrease with their index separation.

\begingroup
\renewcommand{\floatpagefraction}{0.7}
\begin{table}[p]
\centering
\caption{Eight-dimensional validation and test results for checkpoints selected by minimum validation averaged KL during $5{,}000$ training updates.
Validation uses $1{,}024$ generated and $1{,}024$ reference samples; test metrics use $10{,}000$ of each.
KL uses all-point Gaussian estimates with bandwidth $0.03$ and half-weight self contributions.
Outside-sample distances are in normalized coordinates.}
\label{t:eightdim}
\small
\setlength{\tabcolsep}{4pt}
\begin{tabular}{lrrrr}
\toprule
 & \multicolumn{3}{c}{DK training objective} & \\
\cmidrule(lr){2-4}
Metric & Averaged KL & Forward KL & Backward KL & EDM \\
\midrule
Selected update & $3800$ & $3850$ & $3800$ & $4750$ \\
Parameter count & $2148$ & $2148$ & $2148$ & $2270$ \\
Validation averaged KL & $4.86$ & $5.22$ & $4.79$ & $5.50$ \\
Test averaged KL & $2.75$ & $2.90$ & $2.58$ & $2.46$ \\
Test forward KL & $3.32$ & $3.43$ & $3.19$ & $2.27$ \\
Test backward KL & $2.17$ & $2.36$ & $1.96$ & $2.65$ \\
Precision & $96.9\%$ & $94.4\%$ & $97.6\%$ & $86.0\%$ \\
Outside fraction & $3.13\%$ & $5.60\%$ & $2.36\%$ & $14.0\%$ \\
Mean outside distance & $0.0985$ & $0.117$ & $0.0892$ & $0.125$ \\
Outside distance: $95$th percentile & $0.174$ & $0.251$ & $0.117$ & $0.217$ \\
Mode-probability $L^1$ error & $0.128$ & $0.155$ & $0.105$ & $0.0564$ \\
\bottomrule
\end{tabular}
\end{table}

\begin{figure}[p]
\centering
\includegraphics[width=\textwidth]{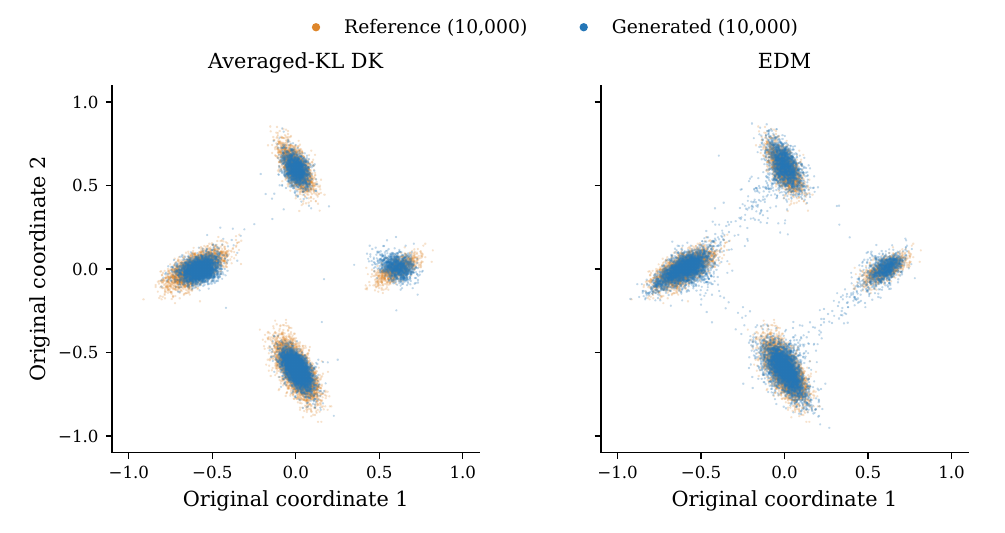}
\caption{Eight-dimensional test samples from the checkpoints selected by minimum validation averaged KL: averaged-KL DK at update $3{,}800$ (left) and plain-MLP EDM at update $4{,}750$ (right).
Each panel shows all $10{,}000$ generated samples in blue and the same $10{,}000$ reference samples in orange.
For visual comparison with the two-dimensional example, we apply $Q_8^\top$ to both normalized sample sets and display the coordinates $U_1/R$ and $U_2/R$.
}
\label{f:eightdim_samples}
\end{figure}

After generating $U$, we apply a fixed orthogonal rotation to obtain the target data:
\[
y^{\mathrm{raw}}=Q_8U.
\]
We construct $Q_8$ once by QR factorization of a standard Gaussian matrix with seed $1729$, and use it for every sample.
We then normalize the rotated data as in \cref{s:density_evaluation}, giving $y=Q_8U/R$.

The eight-dimensional experiments use the training settings in \cref{s:mixture} and the evaluation protocol in \cref{s:density_evaluation}.
The DK drift and diffusion networks retain three hidden layers of width $24$ and two hidden layers of width $18$, respectively, with eight drift outputs and one scalar diffusion output.
We also train $m_0$ and $s_0$, initialized at $0$ and $0.9/\sqrt M$, respectively, for a total of $2{,}148$ DK parameters.
The EDM baseline uses the two-dimensional EDM training and sampling recipe, with three hidden layers of width $26$, eight outputs, and $2{,}270$ trainable parameters.

In the eight-dimensional experiment, DK produces far fewer off-mode samples and smaller mean outside distances than EDM (\cref{t:eightdim,f:eightdim_samples}).
All three DK models have lower validation averaged KL and test backward KL than EDM.
On the test set, EDM has lower forward and averaged KL and better matches the mode probabilities.
These complementary strengths suggest using a score-based generator to capture the modes and their probabilities, followed by DK refinement to reduce off-mode samples.

The eight-dimensional diffusion field is stronger in the gap between the modes and remains close to its lower bound near their centers (\cref{f:diffusion_8d}).
This field is shown at the minimum-validation-KL checkpoint (\cref{f:validation_kl_8d}).

\begin{figure}[tbp]
\centering
\includegraphics[width=\textwidth]{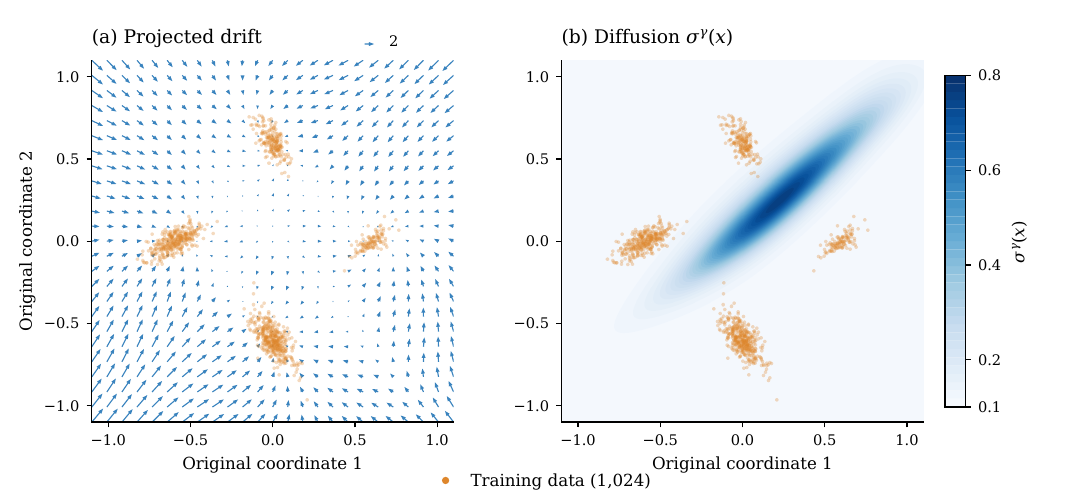}
\caption{Learned eight-dimensional fields at the averaged-KL DK checkpoint selected by minimum validation averaged KL, update $3{,}800$.
Both panels use the normalized coordinates $U_1/R$ and $U_2/R$, obtained by reversing the fixed rotation with $Q_8^\top$, as in \cref{f:eightdim_samples}.
The fields are evaluated on the slice $U_3=\cdots=U_8=0$.
Left: the first two components of $Q_8^\top F^\gamma(x)$.
Right: filled contours of the scalar diffusion.
The arrow scale matches \cref{f:learned_fields}.
The axes match \cref{f:eightdim_samples}, and orange dots show all $1{,}024$ training samples projected onto the displayed coordinate plane.}
\label{f:diffusion_8d}
\end{figure}

As a common reference for DK and EDM, we compare two independent sample sets drawn directly from the target distribution.
This gives estimated averaged KL $2.09$ and precision $90.1\%$.
The DK models exceed this reference in precision, while their estimated averaged KL remains above the reference KL.

The $5{,}000$-update averaged-KL DK and EDM runs take $1.42\times10^3$ and $167$ seconds, respectively.
Their median network-only forward times are $31.2$ and $28.8$ microseconds per batch, respectively.
A more efficient parallel implementation could make DK's two smaller networks faster to evaluate than EDM's single larger network.

\begin{figure}[H]
\centering
\includegraphics[width=\textwidth,trim={0 193.7bp 0 0},clip]{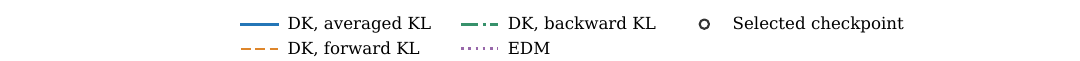}\par
\nointerlineskip
\includegraphics[width=0.5362654\textwidth,trim={0 0 240.4bp 29.5bp},clip]{figures/validation_kl_8d.pdf}
\caption{Validation averaged-KL histories for the eight-dimensional runs, using the same $1{,}024$ reference points, $1{,}024$ generated samples per evaluation, and fixed density bandwidth $0.03$.
The plot shows all $5{,}000$ updates on a logarithmic vertical axis, and markers identify the minimum-validation-KL checkpoints.}
\label{f:validation_kl_8d}
\end{figure}
\endgroup
\section{Conclusion and future work}
\label{s:discussion}

The mixture experiments demonstrate joint optimization of the drift, diffusion, and initial Gaussian parameters using losses that depend directly on the terminal density.
The choice of KL objective changes the fit within the same model, with backward KL giving the smallest mean outside distance in both dimensions.
The learned fields play complementary roles: the drift attracts samples toward the modes, while stronger diffusion helps disperse excess probability from the gaps between them.
Smaller diffusion near the modes limits stochastic spreading.
In each experiment, all three DK models produce fewer off-mode samples and have lower validation averaged KL than the score-based generator.

The density-response formulation supports broader choices of SDE dynamics.
Our experiments use time-independent neural fields for a vector-valued drift and a state-dependent scalar diffusion coefficient; the same response construction also supports time-dependent neural fields.
In either case, the fields can incorporate compatible physical structure or prescribed tail behavior, such as linear attraction.
Extending our generative model to use a general diffusion matrix is another direction for future work.

The DK score formula can also supply training targets for score-based models with more general noising dynamics.
By the DK score identity, the simulated covector $\nu_t$ is a stochastic target for the marginal score, satisfying $\E[\nu_t\mid x_t]=\nabla\log h_t(x_t)$.
With the noising dynamics fixed, a score network $s_\theta(x,t)$ with parameters $\theta$ can be trained by
\[
\min_\theta\ \E\left[\left|s_\theta(x_t,t)-\nu_t\right|^2\right].
\]
This generalizes the analytic conditional-score target construction used in standard denoising score matching \cite{Song2021} by computing targets along the noising paths.

Combining score-based training with DK optimization is a promising direction for future work.
A pretrained score-based generator could initialize the drift, followed by DK optimization of the diffusion and, if needed, refinement of the drift under a selected density-based terminal loss.

\section*{Funding}

This research did not receive any specific grant from funding agencies in the public, commercial, or not-for-profit sectors.

\section*{AI assistance declaration}

The author developed the methodological framework and training algorithms, wrote the core DK implementation, and designed the numerical study.
ChatGPT/Codex assisted with code revisions, supporting scripts, testing, and manuscript preparation under the author's direction.
The author assumes responsibility for all content.

\section*{Data availability statement}

The code for generating the synthetic data and running the numerical experiments is publicly available at \url{https://github.com/niangxiu/DKgen}.

\bibliographystyle{plain}
\bibliography{library}

\end{document}